# Enhancing Pharmacovigilance with Drug Reviews and Social Media

Brent Biseda
University of California, Berkeley
brentbiseda@berkeley.edu

Katie Mo
University of California, Berkeley
kmo@berkeley.edu

## Abstract

This paper explores whether the use of drug reviews and social media could be leveraged as potential alternative sources for pharmacovigilance of adverse drug reactions (ADRs). We examined the performance of BERT (Devlin et al., 2018) alongside two variants that are trained on biomedical papers, BioBERT (Jinhyuk et al., 2019), and clinical notes, Clinical BERT (Alsentzer et al., 2019). A variety of 8 different BERT models were fine-tuned and compared across three different tasks in order to evaluate their relative performance to one another in the ADR tasks. The tasks include sentiment classification of drug reviews, presence of ADR in twitter postings, and named entity recognition of ADRs in twitter postings. BERT demonstrates its flexibility with high performance across all three different pharmacovigilance related tasks.

## 1 Introduction

Pharmacovigilance, the monitoring of potential harmful side effects in medications, plays a critical role in the healthcare system. Drug safety for patients is continually monitored and evaluated after they are released for commercial use. Although pharmaceutical companies evaluate drugs and their adverse drug reactions (ADRs) during clinical trials, the evaluation is limited to a relatively short duration, small sample sizes, and controlled settings. Once the drug is on the market, ADRs are typically reported to the Food and Drug Administration (FDA) through formal reports by patients, healthcare providers, and drug manufacturers. Relying on these mechanisms alone likely results in major underreporting of ADRs.

For any type of medicine, there is an inherent trade-off between its benefit and potential for harm. It is essential for healthcare professionals and patients to be informed and understand the potential risks in order for them to make the best decision regarding the patients' health and well-being. Drug reviews and social media can be leveraged for their quantity and expediency as potential alternative sources that could supplement and enhance the current system of post-market surveillance (Alhuzali and Ananiadou, 2019; Gräßer et al., 2018; Xia et al., 2017; Nikfarjam et al., 2015).

## 2 Background

The field of natural language processing (NLP) has changed dramatically over the last few years. Embeddings from Language Models (ELMo) was introduced and quickly established itself as a breakthrough (Peters et al., 2018). It utilized Long Short Term Memory (LSTM) networks and was capable of creating word representations that utilized the entire sentence context. ELMo outperformed other models across many different natural language processing (NLP) benchmarks. Shortly thereafter, the Bidirectional Encoder Representations from Transformers (BERT) model was created (Devlin et al., 2018), which represented a further breakthrough in the field of NLP. This model has since become the state of the art for a variety of tasks and is created to take context from a sentence both forward and backwards.

Since that time, the BERT model has been adapted to many different areas of research. In this paper, we examine the performance of BERT alongside two variants that are trained on biomedical data, BioBERT (Jinhyuk et al., 2019) and Clinical BERT (Alsentzer et al., 2019). BioBERT was trained on biomedical documents including PubMed abstracts as well as PubMed Central full text articles. Clinical BERT was trained on anonymized patient medical notes and patient discharge summaries. Because the description of ADRs deals specifically with medical terminology, the use of BERT models trained on medical or scientific corpora could result in an improvement when compared to the standard cased and uncased BERT models.

Previous work on sentiment and presence of ADR classification have used N-gram or word2vec approaches to create word vector representations and used Logistic Regression (LR), Naive Bayes (NB), or Support Vector Machines (SVM) as the classification model (Gräßer et al., 2018; Ginn et al., 2014). Other approaches that used deep learning included LSTM or BiLSTM layers followed by a self-attention mechanism (Alhuzali and Ananiadou, 2019). Results from these previous works had an accuracy of 92% for sentiment classification (Gräßer et al., 2018) and F-score of 0.845 for presence of ADR classification from twitter data (Alhuzali and Ananiadou, 2019). Conditional random fields (CRFs) was used in conjunction with K-means clustering on word2vec embeddings in a previous study for named entity recognition of adverse drug reactions and resulted in an F-score of 0.721 on twitter data (Nikfarjam et al., 2015). In this paper, we aim to use the current state of the art techniques such as BERT to see if we can achieve improved performance.

# 3 Methods

We focused on three tasks related to pharmacovigilance which used two different datasets:

1. Sentiment classification using the Drugs.com drug reviews dataset
2. Presence of ADR classification using the Twitter dataset
3. Name entity recognition (NER) detection of ADRs using a subset of the Twitter dataset

A variety of 8 different BERT models were fine-tuned and compared across three different tasks in order to evaluate their relative performance to one another in the ADR tasks. The 8 BERT models consisted of BERT Cased (B-C), BERT Uncased (B-U), BioBERT 1.0 (BB-1.0), BioBERT 1.1 (BB-1.1), Clinical BERT All Notes (CB-A), Clinical BERT Discharge (CB-D), Clinical BioBERT All Notes (CBB-A), and Clinical BioBERT Discharge (CBB-D). The authors of BERT have generally suggested that 2-4 epochs are the range for fine-tuning the model. The default parameters, max sequence length of 128, training batch size of 32, and learning rate of 2e-5, were used.

The code used in this paper is located here: https://github.com/brentbiseda/w266_project

## 3.1 Sentiment Classification

Anonymized user reviews of drugs, their side-effects, and a quantitative user review score were collected from Drugs.com (Gräßer et al., 2018) and is found in the UCI machine learning repository. Our first task is sentiment classification of these user reviews based on the self-reported drug rating. The quantitative score allows for sentiment analysis by aggregating highly positive scores (8 or higher) to serve as a positive review, while highly negative scores are aggregated (3 or lower) and neutral scores (4-7) are aggregated as well. Therefore, we trained our sentiment classifier to discern between these three classes (Table 1).

| Dataset | Positive | Neutral | Negative |
|---|---|---|---|
| Training | 77,907 (60.4%) | 23,114 (17.9%) | 28,017 (21.7%) |
| Dev | 19,503 (60.5%) | 5,710 (17.7%) | 7,046 (21.8%) |
| Test | 32,349 (60.2%) | 9,579 (17.8%) | 11,838 (22.0%) |

*Table 1: Number of examples for the Drugs.com dataset*

We compared the test set accuracy across our set of 8 different BERT variants. Besides the BERT Uncased model, all of the BERT models were trained with the cased configuration. In addition, four different baseline accuracies were established: the most common class, N-gram model with NB classifier, ELMo embeddings with LR, and pre-trained BERT Cased embeddings with LR.

### 3.2 Presence of ADR Classification

The second task is detecting the presence of an ADR within a tweet. The Twitter dataset we used was collected, processed, and annotated by Arizona State University (Ginn et al., 2014). Because we used the Twitter API to retrieve the tweet text, and many of the tweets were no longer available, the dataset that we were able to use only contained 4,169 examples, down from their 10,822 examples. Just like the original dataset, our dataset was heavily skewed towards tweets with no ADR presence. To try to combat the imbalance, we trained models using 3 variants of the training dataset (Table 2). The first training dataset is the original imbalanced dataset. The second training set has the positive class oversampled so that the two classes are balanced. The third training set has the negative class undersampled. The dev and test sets remained the same across the 3 variants.

| Dataset | Im-balanced | Over-sampled | Under-sampled |
|---|---|---|---|
| Training | 2501 (11% positive) | 4440 (50% positive) | 835 (33% positive) |
| Dev | 834 (11% positive) | | |
| Test | 834 (11% positive) | | |

Table 2: Number of examples for the Twitter dataset

For this task, B-C, B-U, and CBB-D models were fine tuned for 3 epochs for each training set. In addition to fine tuning, we extracted features from these models and used them to train an additional classifier. The first [CLS] token embeddings were used to train a LR model. All of the token embeddings were front-padded to the maximum token length and used to train convolutional neural network (CNN) and LSTM models. The most common class and N-gram with NB classifier were used as baseline models.

### 3.3 Named Entity Recognition of ADRs

The third task is detecting the beginning (B), inside (I), and outside (O) of a phrase containing an ADR related mention within a tweet. This task used a subset of the Arizona State University Twitter dataset. The subset consisted of 965 tweets in total, with 500 tweets that contained an ADR and 465 examples that did not contain an ADR. Like any NER task, the large majority of the labels were O tags (Table 3).

| Dataset | B | I | O |
|---|---|---|---|
| Training | 348 (2.8%) | 345 (2.8%) | 11627 (94.4%) |
| Dev | 88 (2.7%) | 93 (2.9%) | 3054 (94.4%) |
| Test | 160 (3.0%) | 117 (2.2%) | 5081 (94.8%) |

Table 3: Number of examples for the NER subset of the Twitter dataset

The baseline models used for this task were the most common class and CRF, which were compared to the fine-tuning of the 8 BERT variants.

## 4 Results and Discussion

### 4.1 Sentiment Classification

The first comparison shown is the sentiment task based on the Drugs.com dataset. We see that the test set accuracy is best with the use of fine-tuning BERT (Table 5). In all cases, each variant of BERT when it is fine-tuned, outperforms each of our baseline evaluation methods (Table 4). We see that BERT embeddings with LR outperformed ELMo embeddings with LR as expected. While the guidance for 2-4 epochs of training was explored, we saw continual performance improvement across all 8 different models when training with 4 epochs. There was no marked performance difference between BERT Cased, BioBERT, or Clinical BERT. However, there was reduced performance with the use of the BERT Uncased model.

| Model | Test Accuracy |
|---|---|
| Most Common Class | 0.602 |
| N-Gram + NB | 0.890 |
| ELMo + LR | 0.709 |
| Pretrained B-C + LR | 0.720 |

*Table 4: Test accuracy of baseline models for sentiment classification*

| Model | 1 Epoch | 2 Epochs | 3 Epochs | 4 Epochs |
|---|---|---|---|---|
| B-C | 0.824 (0.440) | 0.851 (0.421) | 0.876 (0.448) | 0.888 (0.510) |
| B-U | 0.805 (0.483) | 0.820 (0.496) | 0.824 (0.628) | 0.841 (0.737) |
| BB-1.0 | 0.824 (0.444) | 0.854 (0.415) | 0.877 (0.445) | 0.887 (0.504) |
| BB-1.1 | 0.824 (0.442) | 0.854 (0.416) | 0.877 (0.448) | 0.877 (0.492) |
| CB-A | 0.821 (0.445) | 0.854 (0.415) | 0.877 (0.459) | 0.888 (0.515) |
| CB-D | 0.824 (0.444) | 0.855 (0.414) | 0.874 (0.456) | **0.889** (0.529) |
| CBB-A | 0.822 (0.446) | 0.855 (0.419) | 0.873 (0.452) | 0.889 (0.521) |
| CBB-D | 0.823 (0.444) | 0.855 (0.416) | 0.876 (0.460) | 0.888 (0.530) |

*Table 5: Test set accuracy (and loss) from a variety of BERT fine-tuning for sentiment classification*

Because we saw increasing performance in the fine-tuned BERT models from 1-4 epochs, we also explored the effect that additional training could have on the model's accuracy. For the CB-D model, we saw improved performance at 10 epochs with a test accuracy of 0.906 and test loss of 0.693. Due to time and hardware constraints, we did not perform training across all 8 different models.

The CB-D model fine-tuned with 4 epochs had misclassified 5,979 (11.1%) of the reviews. Of those that were misclassified, 5,252 (87.8%) were misclassified from negative or positive sentiment to neutral sentiment, and vice versa. From the remaining misclassifications, 390 (6.5%) reviews were misclassified as positive sentiment when labeled as negative sentiment (false positives), and 337 (5.6%) reviews were misclassified as negative sentiment when labeled as positive sentiment (false negatives). A random sample of 50 false negatives and 50 false positives were analyzed. The majority of false negative reviews, 62%, had mentioned negative side effects, although the user rated the drug highly, while another 22% of reviews had mentioned a different drug than the one being rated. Of the false positive reviews, 40% had mentioned a different drug and 34% had mixed sentiment.

### 4.2 Presence of ADR Classification

A selection of BERT models were fine-tuned with 3 epochs across 3 variants of the training data set (Table 6). We find that fine-tuning B-C, B-U, and CBB-D models were able to improve the F-scores from the baseline N-Gram with NB model. An even larger improvement was seen using BERT extracted features with CNN and LSTM models.

| Model | Im-balanced | Over-sampled | Under-sampled |
|---|---|---|---|
| Most Common | 0 (0.890) | 0 (0.890) | 0 (0.890) |
| N-Gram + NB | 0.197 (0.898) | 0.324 (0.885) | 0.408 (0.888) |
| B-C | 0.570 (0.908) | 0.464 (0.903) | 0.487 (0.854) |
| B-U | 0.590 (0.910) | 0.476 (0.908) | 0.546 (0.841) |

| Model | | | |
|---|---|---|---|
| CBB-D | 0.544 (0.902) | 0.523 (0.912) | 0.510 (0.820) |
| B-U features + LR | 0.562 (0.916) | 0.463 (0.905) | 0.786 (0.950) |
| B-U features + CNN | 0.655 (0.932) | 0.720 (0.945) | 0.951 (0.989) |
| B-U features + LSTM | 0.978 (0.995) | **0.995** (0.999) | **0.995** (0.999) |

*Table 6: Test set F-score (and accuracy) from BERT fine-tuning and models for presence of ADR*

The BERT models seem to handle the imbalanced dataset with some success, as the models fine-tuned with the imbalanced dataset performed better in F-score compared to the oversampled and undersampled datasets. However, using the undersampled training dataset resulted in the best test set F-scores when using BERT extracted features and an additional classifier.

### 4.3 Named Entity Recognition of ADRs

We fine-tuned the 8 different BERT models with 3, 5, and 10 epochs (Table 8) and compared these results to the baseline models (Table 7). We saw improved accuracy with BERT when compared to the two baseline models, with the highest performing model being the B-U model with an F-score of 0.72. This was surprising for an NER task, as typically cased information can be helpful in distinguishing entities. However, tweets do not follow the same formalities as the general written language, and can be frequently seen in all lowercase letters.

| Model | F-score | Accuracy |
|---|---|---|
| Most Common Class | 0.324 | 0.948 |
| CRF | 0.502 | 0.951 |

*Table 7: Test set F-score macro and accuracy of baseline models for NER of ADRs*

| Model | 3 Epochs | 5 Epochs | 10 Epochs |
|---|---|---|---|
| B-C | 0.652 (0.958) | 0.687 (0.960) | 0.687 (0.961) |
| B-U | 0.549 (0.960) | 0.684 (0.960) | **0.720** (0.965) |
| BB-1.0 | 0.489 (0.955) | 0.663 (0.959) | 0.696 (0.962) |
| BB-1.1 | 0.521 (0.953) | 0.661 (0.959) | 0.652 (0.956) |
| CB-A | 0.546 (0.958) | 0.641 (0.951) | 0.662 (0.958) |
| CB-D | 0.546 (0.955) | 0.685 (0.959) | 0.681 (0.961) |
| CBB-A | 0.602 (0.961) | 0.619 (0.955) | 0.646 (0.959) |
| CBB-D | 0.556 (0.958) | 0.647 (0.952) | 0.649 (0.956) |

*Table 8: Test set F-score macro (and accuracy) from a variety of BERT models for NER of ADRs*

Of the words that were predicted as O but labeled as B or I (false negatives), the most frequent words were "of" and "out". In the same vein, the most frequent words being predicted as B or I but labeled as O (false positives) were "my" and "me". Because many of the ADR phrases contain prepositional or possessive words, such as in "sick to my stomach", "messed up my attitude", or "3 days of hell", these words can be difficult for a classifier to distinguish. Notably, many ADRs are not described with any scientific or medical terminology. In some cases, adjectives such as "zombified", "dopey" and "crazy" were used. The presence of colloquial language is reflected in the results from BioBERT and ClincalBERT, as these models actually performed worse compared to regular BERT for this task.

## 5 Conclusion

We can see that while there exist a variety of different BERT variants that have been trained for specialized tasks, their generalizability may be limited when applied to a slightly different domain. While our expectation would have been for superior performance of both BioBERT and Clinical BERT in comparison to regular BERT, we did not see that this was the case. Although pharmacovigilance is within the scientific and medical domain, the 3 tasks we tackled used data generated from patients who may not necessarily use the same language as scientists and healthcare providers.

Regardless of the BERT variant, it appeared that fine-tuning for a larger number of epochs, 5 or 10 epochs in comparison to the recommended 2-4 epochs, had better performance on the test data for sentiment classification and NER. Furthermore, the use of an additional classifier on top of BERT extracted features can provide great benefit, especially when the dataset is limited in size, as shown for ADR classification.

When tackling any related language model task, it can still be worthwhile to explore pre-trained BERT variants to determine an appropriate starting point prior to fine-tuning. BERT demonstrates its flexibility with competitive performance across all 3 different tasks and 2 datasets. Fine-tuning BERT is a powerful tool to be utilized on many natural language processing tasks.